\documentclass[]{interact}

\usepackage{multirow}
\usepackage{epstopdf}
\usepackage[caption=false]{subfig}
\usepackage{pdflscape}
\usepackage{xfrac}
\usepackage{hyperref}

\theoremstyle{plain}

\theoremstyle{definition}

\theoremstyle{remark}

\usepackage[numbers,sort&compress]{natbib}
\bibpunct[, ]{[}{]}{,}{n}{,}{,}

\begin{document}

\articletype{ARTICLE} 
\title{Machine Learning for Two-Sample Testing \\ under Right-Censored Data: A Simulation Study}

\author{
\name{Petr Philonenko\textsuperscript{a}\thanks{CONTACT Petr Philonenko. E-mail: petr-filonenko@mail.ru} and Sergey Postovalov\textsuperscript{b}}
\affil{
    \textsuperscript{a}Sber AI Lab, Moscow, Russia \\
    \textsuperscript{b}Novosibirsk State Technical University, Novosibirsk, Russia
    }
}
\maketitle

\begin{abstract}

The focus of this study is to evaluate the effectiveness of Machine Learning (ML) methods for two-sample testing with right-censored observations. To achieve this, we develop several ML-based methods with varying architectures and implement them as two-sample tests. Each method is an ensemble (stacking) that combines predictions from classical two-sample tests. This paper presents the results of training the proposed ML methods, examines their statistical power compared to classical two-sample tests, analyzes the null distribution of the proposed methods when the null hypothesis is true, and evaluates the significance of the features incorporated into the proposed methods. In total, this work covers 18 methods for two-sample testing under right-censored observations, including the proposed methods and classical well-studied two-sample tests. All results from numerical experiments were obtained from a synthetic dataset generated using the inverse transform sampling method and replicated multiple times through Monte Carlo simulation. To test the two-sample problem with right-censored observations, one can use the proposed two-sample methods (scripts, dataset, and models are available on GitHub and Hugging Face).

\end{abstract}


\section{Introduction}

Two-sample testing (and hypothesis testing in general) on censored data is integral to addressing problems across various fields, including reliability, physics, biology, medicine, economics, bioinformatics, sociology, and engineering~\cite{many_fields}. Recent advanced studies further corroborate this, demonstrating applications of two-sample testing in exploring algorithms in engineering~\cite{intro_engineering}, chemical reactions~\cite{intro_chemical}, new medicines~\cite{intro_medicines}, ecosystems~\cite{intro_ecosystems_1}, and the reliability of engineering systems~\cite{intro_reliability} and dynamic systems~\cite{intro_dynamic}. In other words, two-sample testing is a universal multi-domain approach applicable to a wide array of contemporary scientific challenges.

It is well known that no universally preferred method exists for this problem across all scenarios. Given its wide applicability, there is a constant challenge to improve the tests used. Classical two-sample tests primarily rely on certain parametric or non-parametric assumptions~\cite{lee}, enabling effective differentiation between some competing hypotheses (though they may struggle with others). Consequently, researchers have made numerous attempts to combine and study the properties of different tests~\cite{q_test,combined_tests_1,combined_tests_2}. However, such attempts are complicated by the peculiarities of the subject area, particularly concerning censored (incomplete) observations. Notably, recent works continue to develop robust tests using increasingly complex statistical assumptions and methodologies~\cite{new_two_sample_1}. Machine learning methods, currently applied across various domains~\cite{ml_application}, can also help address this challenge. While there are modern works based on machine learning methods~\cite{juxtapose}, studies often focus more on uncensored observations~\cite{uncensored_case_1}, consider a limited number of methods, and do not compare their effectiveness with classical two-sample tests~\cite{uncensored_case_2}. Additionally, these works frequently pay insufficient attention to the statistical properties of the methods, often limiting their analysis to statistical power~\cite{uncensored_case_3}. It is essential to note that the null distribution can significantly differ under alternatives not considered in prior works, potentially leading to changes in the binary threshold and resulting in incorrect statistical conclusions.

In summary, this is a promising new direction that has not been sufficiently explored from various angles. Therefore, in our work, we consider a wide range of classical two-sample tests applicable to right-censored data that are well-studied in the literature, widely used in practice, and possess distribution-free properties. We explore methods for the two-sample problem using these classical tests as features and forming an ensemble (stacking) for machine learning. The proposed methods are presented and investigated as standard statistical techniques. This approach allows us to combine the advantages of classical tests. For the proposed methods, we investigate power and null distribution under various alternatives using Monte Carlo simulations.

Thus, the main contributions of this study are as follows:
\begin{enumerate}
    \item Develop statistical methods for two-sample testing with right-censored observations using machine learning techniques;
    \item Evaluate the power of the proposed methods in comparison to classical tests;
    \item Study the null distribution numerically under various alternatives to describe the practical applicability of the proposed methods.
    \item Publish repositories of the proposed methods for public use.
\end{enumerate}


\section{Related Works}

As mentioned earlier, existing two-sample tests often rely on constructing a specific sensitive statistic based on certain parametric or non-parametric assumptions. This approach enhances sensitivity to certain types of alternatives.

One group of such tests includes various generalizations of classical tests such as the Cramér–von Mises and Kolmogorov–Smirnov tests for censored data, as shown in the works of Fleming et al.~\cite{FLE80}, Koziol~\cite{KOZ78}, and Schumacher~\cite{SCH84}.

Another group of tests is based on the use of weight functions, such as the weighted log-rank test. The use of different weight functions generates a multitude of different two-sample tests with new properties compared to the original ones. For example, see Gehan~\cite{GEH65}, Peto and Peto~\cite{PET72}, Tarone and Ware~\cite{TAR77}, Prentice~\cite{PRE78}, Kalbfleisch and Prentice~\cite{KAL89}, and Fleming and Harrington~\cite{FLE91}.

The third group of tests is based on the assumption of a specific form of competing hypotheses. This approach was used in the works of Bagdonavičius and Nikulin, who assumed that competing hypotheses could correspond to models with single (SCE-model~\cite{SCE}) or multiple (MCE-model~\cite{MCE}) intersections, as well as the GPH-model~\cite{GPH}.

Additionally, due to their simplicity and efficiency, methods that construct auxiliary functions have gained popularity. These methods involve computing the value of the \textit{selection function} in advance and then using this value to make decisions about the value of the test statistic. For example, Martinez and Naranjo~\cite{q_test}, as well as Philonenko and Postovalov~\cite{MAX}, propose using the selection function to choose between the log-rank and Wilcoxon tests. In another work~\cite{MIN3}, the authors combine the Bagdonavičius-Nikulin and weighted Kaplan-Meier tests using the selection function.

Having established the potential effectiveness of this approach, the question arises of constructing a selection function that is more sensitive to the characteristics of the underlying tests. ML-based methods can be helpful in this regard, as they can be considered a selection function over a set of two-sample tests, using each test as a feature in the model.

Thus, in Section~3, we present the problem formulation, describe classical two-sample tests, and outline the proposed methods for the two-sample problem based on machine learning. In Section~4, we provide the results of numerical experiments, including training results for the proposed methods, a power study comparing the proposed methods to classical tests, an examination of the null distribution, and feature importance results.


\section{Materials \& Methods}
\subsection{Problem Statement}
Suppose that we have two samples of continues variables $\xi_1$ and $\xi_2$ respectively,
$X_1=\{t_{11} , t_{12} , \cdots , t_{1n_1} \}$ and $X_2 = \{t_{21} , t_{22} , \cdots , t_{2n_2} \}$ of two survival distributions $S_1(t)$ and $S_2(t)$. The observation $t_{ij} = min(T_{ij} , C_{ij} )$, where $T_{ij}$ and $C_{ij}$ are the failure and censoring times for the $j$th object of the $i$th group. $T_{ij}$ and $C_{ij}$ are i.i.d. with a cumulative distribution function (CDF) $F_i(t)$ and $F_i^C(t)$ respectively. Survival curve means the probability of survival in the time interval $(0, t)$, i.e. 
$S_i(t) = P\{\xi_i > t\} = 1 - F_i(t).$

Then the \textbf{null hypothesis} is
$H_0 : S_1(t) = S_2(t)$
against an \textbf{alternative hypothesis}
$H_A: S_1(t) \neq S_2(t)$.

Further, we suppose that the elements of samples are ordered: $t_{11} < \cdots < t_{1n_1}$ and $t_{21} < \cdots < t_{2n_2}$.
Denote \textit{the censoring indicators} $c_{ij}$ as 
$c_{ij}~=~
\begin{cases}
0, t_{ij} \text{~is failure,} \\
1, t_{ij} \text{~is censored,} \\
\end{cases}$
and \textit{the censoring rate} $r_{i}$ as 
$r_i = \sum_{j=1}^{n_i} c_{ij} / n_i$.

\subsection{Two-Sample Tests}

\subsubsection{Log-rank test}
The \textbf{log-rank test} is also called as \textit{Mantel–Haenszel test} or \textit{Mantel–Cox test}. The test statistic \cite{log_rank,mantel,lee} is
\[S_{LG} = \frac{ \sum_{i=1}^n (1-c_i) - \sum_{j=1}^i \frac{1}{n-j+1} }{ \sqrt{ (\sum_{i=1}^n (1-c_i) \frac{n-i}{n-i+1}) \frac{n_1n_2}{n(n-1)}}   }.\]

The null hypothesis $H_0$ is rejected with the significance level $\alpha$, if $\left| S_{LG} \right| > z_{1-\alpha/2}$, where $z_{1-\alpha/2}$ is the $(1-\alpha/2)$-quantile of the standard normal distribution.

\subsubsection{Generalizations of Wilcoxon test}

\textbf{\textit{Gehan’s generalized Wilcoxon test.}} This test is also called as \textit{Gehan-Breslow-Wilcoxon test}. The test statistic \cite{lee,GEH65} is
\[S_G = \frac{ \sum_{i=1}^n (1-\nu_i)h_i }{ \sqrt{ \frac{n_1n_2}{n(n-1)} \sum_{i-1}^n (1-\nu_i)h_i^2 } },\]
where
$h_i = \sum_{j=1}^n \nu_j h_{ij}$
and
$h_{ij} = 
\begin{cases}
+1, t_i > t_j ~\&~ c_j=0 ~\&~ \nu_i=0 ~\&~ \nu_j=1, \\
-1, t_i < t_j ~\&~ c_j=0 ~\&~ \nu_i=0 ~\&~ \nu_j=1, \\
0, \text{~otherwise.} \\
\end{cases}$

The null hypothesis $H_0$ is rejected with the significance level $\alpha$, if $\left| S_G \right| > z_{1-\alpha/2}$, where $z_{1-\alpha/2}$ is the $(1-\alpha/2)$-quantile of the standard normal distribution.

\textit{\textbf{Peto and Peto’s generalized Wilcoxon test.}} The test statistic \cite{PET72,lee} is
\[S_P = \frac{ n(n-1) \sum_{i=1}^n u_i(1-\nu_i) }{n_1 n_2 \sum_{i=1}^{n} u_i^2},\]
where
$u_i = 
\begin{cases}
\hat{S}_{KM}(t_i) + \hat{S}_{KM}(t_{k_i}), c=0, \\ 
\hat{S}_{KM}(t_{k_i}) - 1, c=1, \\
\end{cases}$
$k_i = max\{ j | j \in \{0, ..., i-1\}, c_j=0 \}$, $c_0=0$
and $\hat{S}_{KM}(t)$ is the Kaplan-Meier~\cite{kaplan_meier} estimator of $S(t)$ by the pooled sample~$T$.

The null hypothesis $H_0$ is rejected with the significance level $\alpha$, if $\left| S_P \right| > z_{1-\alpha/2}$, where $z_{1-\alpha/2}$ is the $(1-\alpha/2)$-quantile of the standard normal distribution.

\subsubsection{Weighted tests}

\textbf{\textit{Weighted log-rank test.}}
Based on the findings in~\cite{SCE}, we describe the statistics in a unified format, covered all previously presented tests and incorporating additional ones. The test statistic can be computed as
$S_{WLG} = U^2 / \sigma$
where
\[U = \sum_{j=1}^{n_1} (1 - c_{1j})K(t_{1j})\frac{Y_2(t_{1j})}{Y(t_{1j})}   + \sum_{j=1}^{n_2} (1 - c_{2j})K(t_{2j})\frac{Y_1(t_{2j})}{Y(t_{2j})},\]
\[\sigma = \sum_{i=1}^{2} \sum_{j=1}^{n_i} (1 - c_{ij}) K^2(t_{ij}) 
\frac{ Y_1(t_{ij})Y_2(t_{ij}) }{ Y^2(t_{ij}) },\]
\[Y(t) = Y_1(t) + Y_2(t),  Y_i(t) = \sum_{j=1}^{n_i}Y_{ij}(t), Y_{ij} = 1_{t_{ij} \ge t}, i=1,2.\]

Using a certain parameterizations of $K(t)$, it is possible to vary a behaviour of the test statistic. We present some possible examples of parameterization:
\begin{enumerate}
    \item \textit{log-rank (Mantel–Haenszel) statistic}: $K(t) = 1 / \sqrt{n_1 + n_2}$;

    \item \textit{Gehan-Breslow-Wilcoxon statistic}: $K(t) = Y(t) / \sqrt{(n_1 + n_2)^3}$;

    \item \textit{Peto-Peto-Prentice statistic}: $K(t) = \hat{S}(t) / \sqrt{n_1 + n_2}$;

    \item \textit{Tarone-Ware statistic}: $K(t) = \sqrt{Y(t)} / (n_1 + n_2)$;
        
    \item \textit{Prentice statistic}: $K(t) = \frac{\hat{S}(t)}{\sqrt{n_1+n_2}} \cdot \frac{Y(t)}{Y(t)+1}$;
\end{enumerate}
where $\hat{S}(t)$ is the Kaplan-Meier estimator on the pooled sample.

The test statistic $S_{WLG}$ is asymptotically distributed as chi-square distribution with one degree of freedom. The test statistic has a right-critical area.

\textbf{\textit{Weighted Kaplan-Meier test.}} The weighted Kaplan–Meier test statistic measures the average difference between the survival functions.

Let $\hat{S}_1(t)$ and $\hat{S}_2(t)$ are the Kaplan–Meier estimates of the $S_1(t)$ and $S_2(t)$ consequently, and $\hat{S}(t)$ is the Kaplan–Meier estimate of the pooled sample. For every sample $X_i$, we construct additional sample $\overline{X}_i$ changing the indicator $c_{ij}$ to the opposite $\overline{c}_{ij} = 1 - c_{ij}$. Thus, $\hat{S}_1^C(t)$ and $\hat{S}^C_2(t)$ are the Kaplan–Meier estimates for additional samples $\overline{X}_1$ and $\overline{X}_2$.

The test statistic~\cite{WKM} can be computed as $S_{WKM} = U / \sqrt{\sigma^2}$
where
\[U = \sqrt{ \frac{n_1 n_2}{n_1+n_2} } \int_{0}^{T_C} \hat{w}(t) \{ \hat{S_2}(t) - \hat{S_1}(t) \} dt, ~
\hat{w}(t) = \frac{ (n_1+n_2) \hat{S_1}^C(t-) \hat{S_2}^C(t-) }{ n_1 \hat{S_1}^C(t-) + n_2 \hat{S_2}^C(t-) } \]
and
$ T_C = \max_{t \in R} \left( \min \{ \hat{S_1}(t), \hat{S_2}(t), \hat{S_1}^C(t), \hat{S_2}^C(t) \} > 0 \right). $

The weight function $w(t)$ downweights the contributions of the $S_2(t) - S_1(t)$ in $U$ over later time periods if censoring is heavy. The statistic $U$ is asymptotically normal distributed with a variance $\sigma^2$:

\[ \sigma^2 = - \int_{0}^{T_C} \left( \int_{t}^{T_C} 
\hat{w}(\tau)\hat{S}(\tau)d\tau \right)^2 
\frac{ n_1\hat{S}_1^C(t-) + n_2\hat{S}_2^C(t-) }{ (n_1+n_2) \left( \hat{S}_1^C(t-)\hat{S}_2^C(t-) \right) }
\frac{ d(\hat{S}(t)-\hat{S}(t-)) }{ \hat{S}(t)\hat{S}(t-) }.
\]

The test statistic $S_{WKM}$ is asymptotically distributed as the standard normal distribution $N(0, 1)$ with two-side critical area.

\subsubsection{Bagdonavičius-Nikulin tests}
\textbf{\textit{Test based on the Simple Cross-Effect (SCE) model.}} The test statistic~\cite{SCE} is 
$S_{\text{BN-SCE}} = (U_1,U_2)^T \Sigma^{-1} (U_1,U_2)$
where
\[U_1 = \sum_{j=1}^{n_1} (1-c_{1j}) \frac{K_1(t_{1j})}{Y(t_{1j})}Y_2(t_{1j}) - \sum_{j=1}^{n_2} (1-c_{2j}) \frac{K_1(t_{2j})}{Y(t_{2j})}Y_1(t_{2j}),\]
\[U_2 = \sum_{j=1}^{n_1} (1-c_{1j}) \frac{K_2(t_{1j})}{Y(t_{1j})}Y_2(t_{1j}) - \sum_{j=1}^{n_2} (1-c_{2j}) \frac{K_2(t_{2j})}{Y(t_{2j})}Y_1(t_{2j}).\]

The elements of the matrix $\Sigma$ can be computed as
\[\sigma_{ij} = \sum_{r=1}^{2}\sum_{s=1}^{n_r} (1-c_{rs})K_i(t_{rs}) \frac{Y_1(t_{rs})Y_2(t_{rs})}{Y^2(t_{rs})}.\]

The following functions are $K_1(t)$ and $K_2(t)$ may be used
\[K_1(t) = exp(-\Lambda(t))/\sqrt{n}, K_2(t) = (exp(-\Lambda(t))-1)/\sqrt{n},\]
\[Y(t) = Y_1(t) + Y_2(t), Y_i(t) = \sum_{j=1}^{n_i}Y_{ij}(t), Y_{ij}(t)=1_{t_{ij} \ge t}, i=1,2, \] 
\[\Lambda(t) = \sum_{i=1}^2 \sum_{j: c_{ij}-0, t_{ij}\le t} \frac{1}{Y(t_{ij})}.\]

The test statistic $S_{\text{BN-SCE}}$ is asymptotically distributed as chi-square distribution with two degrees of freedom. The test has right-side critical area.

\textbf{\textit{Test based on the Multiple Cross-Effect (MCE) model.}} The test statistic~\cite{MCE} is 
$S_{\text{BN-MCE}} = (U_1,U_2,U_3)^T \Sigma^{-1} (U_1,U_2,U_3),$
where
\[U_1 = \sum_{ j: c_{1j}=0 } \frac{Y_2(t_{1j})}{Y(t_{1j})} - \sum_{ j: c_{2j}=0 } \frac{Y_1(t_{2j})}{Y(t_{2j})},
U_2 = - \sum_{ j: c_{1j}=0 } \frac{Y_2(t_{1j})}{Y(t_{1j})} \Lambda(t_{1j}) - \sum_{ j: c_{2j}=0 } \frac{Y_1(t_{2j})}{Y(t_{2j})} \Lambda(t_{2j}),\]
\[U_3 = - \sum_{ j: c_{1j}=0 } \frac{Y_2(t_{1j})}{Y(t_{1j})} \Lambda^2(t_{1j}) - \sum_{ j: c_{2j}=0 } \frac{Y_1(t_{2j})}{Y(t_{2j})} \Lambda^2(t_{2j}).\]

The elements of the matrix $\Sigma$ can be computed as
\[\sigma_{11} = \sum_{i=1}^{2} \sum_{j:c_{ij}=0} \frac{Y_1(t_{ij})Y_2(t_{ij})}{Y^2(t_{ij})},
\sigma_{12} = \sigma_{21} = \sum_{i=1}^{2} \sum_{j:c_{ij}=0} \frac{Y_1(t_{ij})Y_2(t_{ij})}{Y^2(t_{ij})} \Lambda(t_{ij}),\]
\[\sigma_{13} = \sigma_{31} = \sigma_{22} = \sum_{i=1}^{2} \sum_{j:c_{ij}=0} \frac{Y_1(t_{ij})Y_2(t_{ij})}{Y^2(t_{ij})} \Lambda^2(t_{ij}),\]
\[\sigma_{23} = \sigma_{32} = \sum_{i=1}^{2} \sum_{j:c_{ij}=0} \frac{Y_1(t_{ij})Y_2(t_{ij})}{Y^2(t_{ij})} \Lambda^3(t_{ij}),
\sigma_{33} = \sum_{i=1}^{2} \sum_{j:c_{ij}=0} \frac{Y_1(t_{ij})Y_2(t_{ij})}{Y^2(t_{ij})} \Lambda^4(t_{ij}).\]

The test statistic $S_{\text{BN-MCE}}$ is asymptotically distributed as chi-square distribution with three degrees of freedom. The test has right-side critical area.

\textbf{\textit{Test based on the Generalized Proportional Hazards (GPH) model.}} The test statistic~\cite{GPH} is 
$S_{\text{BN-GPH}} = (U_1,U_2)^T \Sigma^{-1} (U_1,U_2)$
where
\[U_1 = \sum_{j: c_{1j}=0} \frac{Y_2(t_{1j})}{Y(t_{1j})} - \sum_{j: c_{2j}=0} \frac{Y_1(t_{2j})}{Y(t_{2j})},\]
\[U_2 = - \sum_{j: c_{1j}=0} \frac{Y_2(t_{1j})}{Y(t_{1j})} \ln(1+\Lambda(t_{1j})) + \sum_{j: c_{2j}=0} \frac{Y_1(t_{2j})}{Y(t_{2j})} \ln(1+\Lambda(t_{2j})).\]

The elements of the matrix $\Sigma$ can be computed as
\[\sigma_{11} = \sum_{i=1}^{2} \sum_{j: c_{ij}=0} Y_1(t_{ij})Y_2(t_{ij}) / Y^2(t_{ij}),\]
\[\sigma_{22} = \sum_{i=1}^{2} \sum_{j: c_{ij}=0} Y_1(t_{ij})Y_2(t_{ij}) \ln^2(1+\Lambda(t_{ij})) / Y^2(t_{ij}),\]
\[\sigma_{12} = \sigma_{21} = \sum_{i=1}^{2} \sum_{j: c_{ij}=0} Y_1(t_{ij})Y_2(t_{ij}) \ln(1+\Lambda(t_{ij})) / Y^2(t_{ij}).\]

The test statistic $S_{\text{BN-GPH}}$ is asymptotically distributed as chi-square distribution with two degrees of freedom. The test has right-side critical area.

\subsubsection{Two-Stage tests}

\textbf{Q-test.} The test statistic~\cite{q_test} is $S_Q = S_{LG}$ if $Q < 0$ else $S_{P}$ where $Q = \{ \hat{S_2}(q_{0.6}) - \hat{S_1}(q_{0.6}) \} - \{ \hat{S_2}(q_{0.2}) - \hat{S_1}(q_{0.2}) \}$, $q_p = \hat{S_1}^{-1}(p)$, and $\hat{S_i}(t)$ is the Kaplan-Meier estimator. 

The test statistic $S_Q$ has a double-side critical area and the $H_0$ is rejected under a large value of the $|S_Q|$.

\textbf{\textit{Maximum value test.}} The test statistic~\cite{MAX} is 
$S_{MAX} = \max (|S_{LG}|, |S_{G}|)$. 

The test statistic $S_{MAX}$ has right-side critical area. The distribution of the test statistic $S_{MAX}$ under the $H_0$ is represented in~\cite{MAX_distr}.

\textbf{\textit{MIN3 test.}} The test statistic~\cite{MIN3} is 
$S_{MIN3} = \min( pv_{WKM}, pv_{\text{BN-MCE}}, pv_{\text{BN-GPH}})$
where
$pv_{WKM} = 2 \cdot \min\{ F_{N(0,1)}(S_{WKM}), 1 - F_{N(0,1)}(S_{WKM}) \},$
$pv_{\text{BN-MCE}} = 1 - F_{\chi^2(3)}(S_{\text{BN-MCE}})$,
$pv_{\text{BN-GPH}} = 1 - F_{\chi^2(2)}(S_{\text{BN-GPH}})$.

The test statistic $S_{MIN3}$ has left-side critical area. The distribution of the test statistic $S_{MIN3}$ under the $H_0$ is represented in~\cite{MIN3}.

\subsection{Proposed ML-based Methods for Two Sample Problem}

The two-sample tests described earlier are classical methods for addressing this problem. These tests offer several advantages: 
(1) They are well-studied, with extensive theoretical investigations available.
(2) They have been used for many years to solve applied problems (e.g., the log-rank test has been utilized for decades in medical research).
(3) Almost all tests possess the important property of being distribution-free, ensuring asymptotic convergence to the corresponding limit distribution.
(4) Each test is preferable under specific alternative hypotheses, where other tests may be unsuitable.

This suggests that combining these statistics can yield a test statistic that is more comprehensive and sensitive to a broader range of alternatives than each individual test statistic. Furthermore, such a combined test would be more robust and powerful. To achieve this, we consider these tests as weak classifiers that can be integrated into a strong classifier.

\begin{figure}[]
    \centering
    \includegraphics[width=\textwidth]{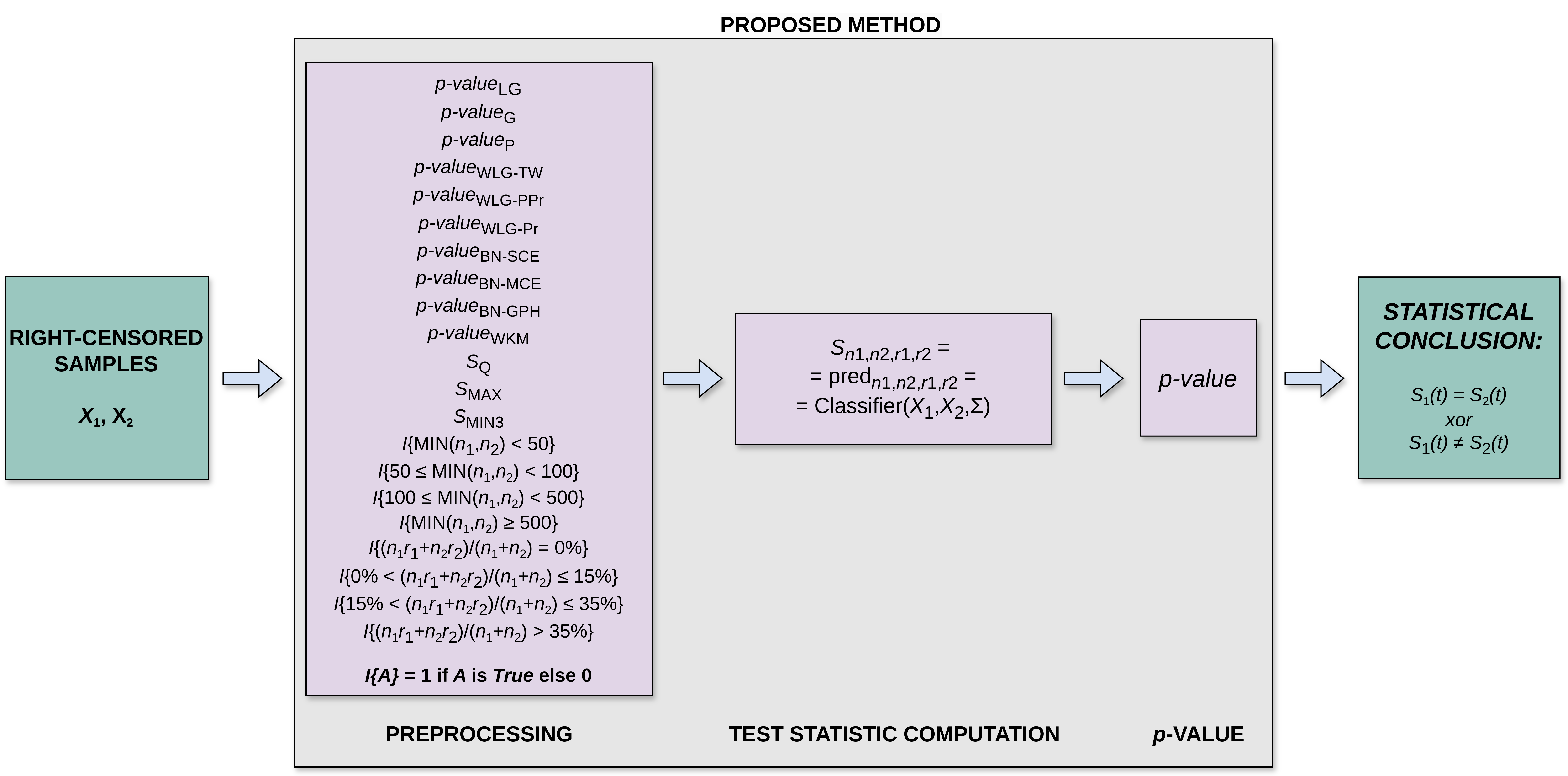}
    \caption{Implementation flow chart of the proposed methods}
    \label{fig:ml_flow}
\end{figure}

We formulate the two-sample problem as a \textbf{binary classification} task: \textbf{0 if \(H_0\) is accepted, 1 if \(H_0\) is rejected}. Then, the classifier's input consists of two right-censored samples, while the output is a predicted logit in [0, 1], representing the "probability" of rejecting \(H_0\). Each two-sample test is treated as an independent feature, allowing us to combine the properties of various statistical methods using a stacking architecture to solve the problem. Thus, a statistical conclusion \(D\) through the proposed methods can be formulated as follows:
\[
D = 
\begin{cases}
    H_0~\text{is accepted, if}~p_v \ge \alpha \\
    H_0~\text{is rejected, if}~p_v < \alpha
\end{cases}
\]
where
\(p_v = 1 - G_{n_1,n_2,r_1,r_2}(pred_{n_1,n_2,r_1,r_2}|H_0)\), 
\(pred_{n_1, n_2, r_1, r_2} = \text{Classifier}(X_1, X_2, \Sigma)\), 
\(G_{n_1,n_2,r_1,r_2}(\cdot|H_0)\) is the pre-limit null distribution,
\(\alpha\) is the probability of a Type I error,
\(\Sigma = \Sigma_1 \cup \Sigma_2\), 
\(\Sigma_1 = \{ S_P, S_G, S_{LG}, S_{\text{SCE}}, S_{\text{MCE}}, S_{\text{GPH}}, S_{\text{TW}}, S_{\text{PPr}}, S_{\text{Pr}}, S_{WKM} \}\), 
and \(\Sigma_2 = \{ S_Q, S_{MAX}, S_{MIN3} \}\). We also detailly demonstrate the process of the described method in Figure~\ref{fig:ml_flow}. Hence, the test statistic for each proposed method \(S_{\text{ML}}\) is given by \(S_{\text{ML}} = pred_{n_1,n_2,r_1,r_2}\). This representation allows for the investigation of all standard statistical procedures (e.g., test power and null distribution).

A binary Classifier$(X_1,X_2,\Sigma)$ needs to be trained on the target $y\in\{0,1\}$ ($y = 0$ if the $H_0$ is satisfied for $X_1, X_2$; otherwise, $y = 1$) and on features that are formed as follows: 
\[
\text{Classifier}(X_1,X_2,\Sigma) :=
\begin{cases}
    p\text{-value}(X_1, X_2, S_l), & \text{if } S_l \in \Sigma_1, \\
    S_l(X_1, X_2), & \text{if } S_l \in \Sigma_2, \\
    I( \min(n_1, n_2) \in [n_L, n_R]), \\
    I( \frac{n_1r_1 + n_2r_2}{n_1+n_2} \in [r_L, r_R]),
\end{cases}
\]
where \(I(A) = 1\) if \(A\) is True; otherwise, \(I(A) = 0\). We set \([n_L, n_R]\) to represent small 50-, low (50, 100], medium (100, 500], and high 500+ sample sizes. For \([r_L, r_R]\), we categorize it as null [0\%, 0\%], low (0\%, 15\%], medium (15\%, 35\%], and high 35\%+ censoring rates. For the proposed methods, we limit the number of features during the feature engineering stage since the goal is to enhance information about the properties of the two-sample tests rather than overfitting on the selected alternatives.

To solve the binary classification task, we apply several machine learning methods:
(1) \textit{Logistic Regression} (denoted as \textit{LogReg}) implemented in sklearn;
(2) \textit{Random Forest} (denoted as \textit{RF}) implemented in sklearn;
(3) \textit{Gradient Boosting Machine} (denoted as \textit{GBM}) implemented in sklearn;
(4) \textit{XGBoost};
(5) \textit{CatBoost} proposed by Yandex~\cite{catboost};
(6) \textit{LightAutoML} (\textit{LAMA}) proposed by Sber AI Lab~\cite{lama}.

For the proposed methods, the limit distribution $G(x|H_0)$ is unknown in analytical form. However, this is not necessary for applying the statistical method, as it is sufficient to study how the pre-limit distribution $G_{n_1,n_2,r_1,r_2}(x|H_0)$ can significantly vary. Therefore, we numerically study the pre-limit distribution of the proposed methods and determine the possibility of their applicability in practice. To strengthen the conviction of the numerical results obtained, we use competing hypotheses of diverse distribution families $F(t)$ and $F^C(t)$, as described in Section~\ref{Alt_Hyp}. We also note that the proposed methods have a \textit{right-tailed critical area} (according to the problem statement based on binary classification) and the null hypothesis $H_0$ is rejected when the test statistic is large.

In training ML models, we select the best models by maximizing the Accuracy metric. Since we have a balanced dataset (50\% where $H_0$ is true and 50\% where $H_0$ is rejected) of synthetic data, maximizing Accuracy allows us to minimize the probabilities of Type I and Type II errors. The synthetic data were simulated using Monte Carlo replications. Additionally, we report other metrics for reference: Precision, Recall, ROC AUC, Average Precision, Specificity, and Proportion of predictions without FN (false negative): $(TP+FP+TN) / (TP+FP+FN+TN)$.

The proposed methods can also be used to solve the two-sample problem and are available on \href{https://github.com/pfilonenko/ML_for_TwoSampleTesting}{GitHub}.

\subsection{Alternative Hypotheses} \label{Alt_Hyp}

We study the performance of two-sample tests using alternative hypotheses where each alternative is a pair of competing hypotheses $S_1(t)$ and $S_2(t)$. Studies~\cite{combined_tests_1,combined_tests_2} examine the impact of alternatives with differences at early, middle, and late time on the test power. Studies of Bagdonavičius and Nikulin~\cite{SCE,MCE} use another way based on alternatives with different numbers of intersections of competing hypotheses. We combine these experiments together and provide experiments that have alternatives with 0, 1, and 2 intersections at early, middle, and late time. 

\begin{figure}[ht]
    \centering
    \includegraphics[width=7.5cm]{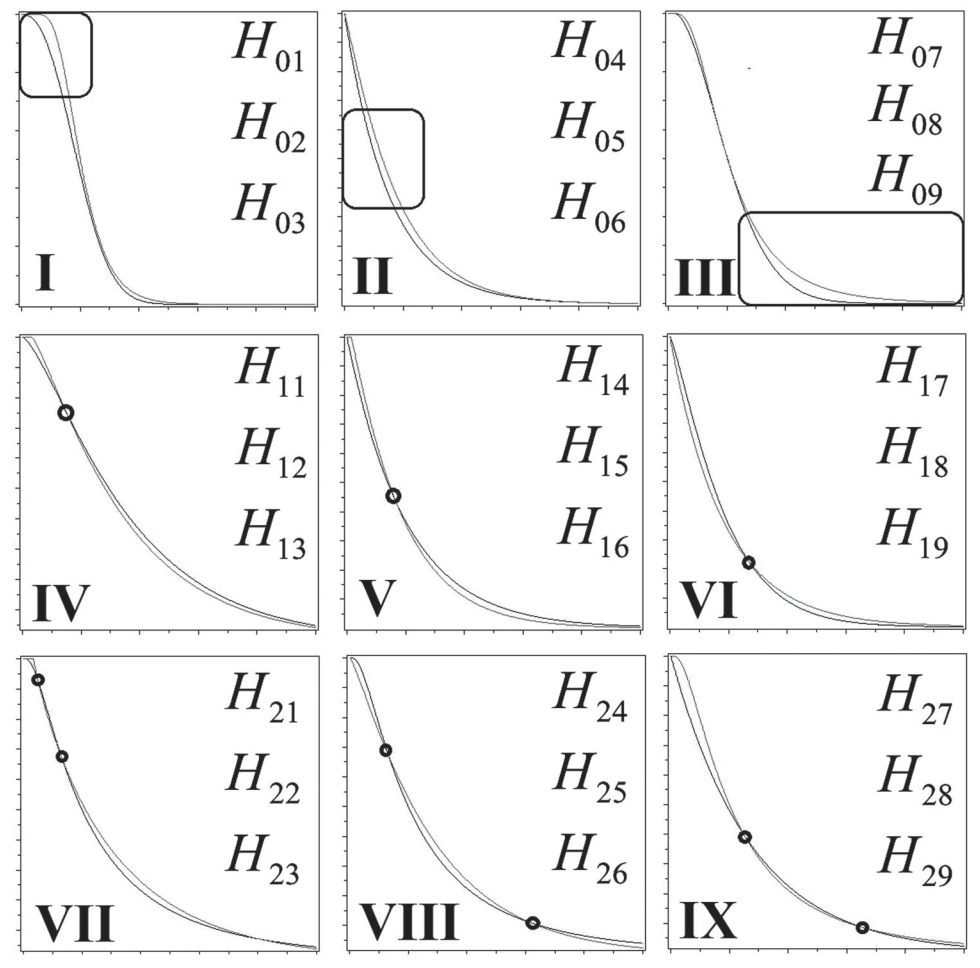}
    \caption{Groups of alternative hypotheses}
    \label{fig:alt_hip}
\end{figure}

Taking into account that the findings obtained for one alternative are not representative in the general case. Therefore, we form \textit{groups of similar alternatives} that contain various closest competing hypotheses. The \textit{alternative groups} are represented in Figure~\ref{fig:alt_hip}. Additionally, in order to reduce the impact of a random choice for alternative, we vary families of $F(t)$ and $F^C(t)$ for each alternative. The following families of distributions are used in the construction of alternative hypotheses:
\begin{enumerate}
    \item $f_{Exp} (x; \mu, \lambda) = \lambda \exp ( -\lambda(x - \mu) ), x \ge \mu$;
    \item $f_{We} (x; \mu, \lambda, \nu) = \nu (x - \mu)^{\nu-1} \exp( -((x-\mu)/\lambda)^{\nu})/\lambda^{\nu}, x \ge \mu$;
    \item $f_{G} (x; \mu, \lambda, \nu) = ((x - \mu)/\lambda)^{\nu-1} \exp( -(x-\mu)/\lambda ) \lambda \Gamma(\nu), x \ge \mu$;
    \item $f_{LgN} (x; \mu, \lambda) = \exp( -(\ln{x} - \mu)^2 / 2\lambda^2) / x \sqrt{2\pi\lambda^2}       , x \ge 0$.
\end{enumerate}

Note that we fix the size of inequality between competing hypotheses $S_1(t)$ and $S_2(t)$ by $L_1 = 0.1$ where $L_1$ is the $l1$-norm. It is necessary to ensure the closest experimental conditions between alternatives. The choice of this value is due to the following reasons:
\begin{enumerate}
    \item It provides similar conditions between alternatives to ensure numerical experiments.
    \item It provides the test power $1-\beta$ to be in the range $\alpha < 1-\beta < 1$ for most considered sample sizes where $\alpha$ and $\beta$ are errors of the I and II types.
\end{enumerate}

\textit{Censoring rate} $r$ can be analytically computed as 
$r = \int_{-\infty}^{+\infty} \left( 1 - F^C(t;\theta) \right) f(t;\eta){dt}$ 
where 
$0 \le r \le 1$, 
$f(t;\eta) = dF(t;\eta) / dt$ and $\eta$, $\theta$ 
are some parameters of the 
$F(t;\eta)$ and $F^C(t;\theta)$.
If we fix $\eta$ and $r$, then we can find the parameters $\theta$ of the $F^C(t;\theta)$ by solving the corresponding equation above. This is exactly how alternatives were constructed in this study. As a result, we present the parameters of alternatives in Table~\ref{tab:alt_hip}.

\begin{table}[ht]
\centering
\caption{Groups of alternative hypotheses}
\label{tab:alt_hip}
\begin{tabular}{ccccc}
\hline
\textbf{Group}        & \textbf{\begin{tabular}[c]{@{}c@{}}Description\end{tabular}}                                             & \textbf{$H_i$} & \textbf{$S_1(t)$}        & \textbf{$S_2(t)$}          \\ \hline
\multirow{3}{*}{I}    & \multirow{3}{*}{\begin{tabular}[c]{@{}c@{}}0 intersections \\ with diﬀerence \\ in the early time\end{tabular}}  & $H_{01}$       & $Exp(0, 1)$              & $ Exp(0.1, 1)$             \\
                      &                                                                                                                  & $H_{02}$       & $We(0, 1.1, 2.4)$        & $ LgN(0, 0.370)$           \\
                      &                                                                                                                  & $H_{03}$       & $LgN(0.01, 0.913)$       & $ Exp(0, 0.742)$           \\ \hline
\multirow{3}{*}{II}   & \multirow{3}{*}{\begin{tabular}[c]{@{}c@{}}0 intersections \\ with diﬀerence \\ in the middle time\end{tabular}} & $H_{04}$       & $G(0, 1.060, 1.160)$     & $ Exp(0, 0.863)$           \\
                      &                                                                                                                  & $H_{05}$       & $Exp(0, 1.3)$            & $ We(0, 0.9, 1.1)$         \\
                      &                                                                                                                  & $H_{06}$       & $Exp(0, 1)$              & $ We(0.09, 1.1, 1.07)$     \\ \hline
\multirow{3}{*}{III}  & \multirow{3}{*}{\begin{tabular}[c]{@{}c@{}}0 intersections \\ with diﬀerence \\ in the late time\end{tabular}}   & $H_{07}$       & $Exp(0, 1.3)$            & $ G(0, 0.806, 1.064)$      \\
                      &                                                                                                                  & $H_{08}$       & $We(0.5, 1, 1.2)$        & $ Exp(0.567, 1)$           \\
                      &                                                                                                                  & $H_{09}$       & $We(0.118, 1.1, 1.735)$  & $ LgN(0.01, 0.6)$          \\ \hline
\multirow{3}{*}{IV}   & \multirow{3}{*}{\begin{tabular}[c]{@{}c@{}}1 intersection \\ in the \\ early time\end{tabular}}                  & $H_{11}$       & $Exp(0, 1)$              & $ Exp(0.05, 1.159)$        \\
                      &                                                                                                                  & $H_{12}$       & $G(0, 1.273, 1.475)$     & $ G(0.159, 1.300, 1.273)$  \\
                      &                                                                                                                  & $H_{13}$       & $We(0.02, 1, 1.1)$       & $ Exp(0, 0.909)$           \\ \hline
\multirow{3}{*}{V}    & \multirow{3}{*}{\begin{tabular}[c]{@{}c@{}}1 intersection \\ in the \\ middle time\end{tabular}}                 & $H_{14}$       & $We(0, 0.980, 0.905)$    & $ G(0, 0.972, 0.974)$      \\
                      &                                                                                                                  & $H_{15}$       & $Exp(0, 1)$              & $ We(0.071, 0.906, 1.059)$ \\
                      &                                                                                                                  & $H_{16}$       & $G(0.01, 1, 1.15)$       & $ Exp(0, 0.833)$           \\ \hline
\multirow{3}{*}{VI}   & \multirow{3}{*}{\begin{tabular}[c]{@{}c@{}}1 intersection \\ in the \\ late time\end{tabular}}                   & $H_{17}$       & $We(0, 0.968, 1.214)$    & $ Exp(0, 1.107)$           \\
                      &                                                                                                                  & $H_{18}$       & $G(0, 1.1, 1.040)$       & $ G(0, 0.9, 1.302)$        \\
                      &                                                                                                                  & $H_{19}$       & $We(0.5, 1.1, 1.1)$      & $ Exp(0.471, 1)$           \\ \hline
\multirow{3}{*}{VII}  & \multirow{3}{*}{\begin{tabular}[c]{@{}c@{}}2 intersections \\ in the early \\ and middle time\end{tabular}}      & $H_{21}$       & $LgN(0, 0.948)$          & $ We(0.173, 1.325, 0.911)$ \\
                      &                                                                                                                  & $H_{22}$       & $Exp(0.5, 1.047)$        & $ LgN(0.141, 0.596)$       \\
                      &                                                                                                                  & $H_{23}$       & $We(0.5, 1, 1.2)$        & $ Exp(0.530, 1)$           \\ \hline
\multirow{3}{*}{VIII} & \multirow{3}{*}{\begin{tabular}[c]{@{}c@{}}2 intersections \\ in the early \\ and late time\end{tabular}}        & $H_{24}$       & $LgN(0, 0.916)$          & $ G(0.01, 1.213, 1.192)$   \\
                      &                                                                                                                  & $H_{25}$       & $LgN(0, 0.817)$          & $ Exp(0.185, 0.818)$       \\
                      &                                                                                                                  & $H_{26}$       & $We(0.01, 1.697, 1.846)$ & $ LgN(0.293, 0.569)$       \\ \hline
\multirow{3}{*}{IX}   & \multirow{3}{*}{\begin{tabular}[c]{@{}c@{}}2 intersections \\ in the middle \\ and late time\end{tabular}}       & $H_{27}$       & $We(0, 1.355, 1.018)$    & $ LgN(0.000, 0.867)$       \\
                      &                                                                                                                  & $H_{28}$       & $G(0, 1.134, 1.231)$     & $ LgN(0, 0.876)$           \\
                      &                                                                                                                  & $H_{29}$       & $Exp(0, 0.744)$          & $ LgN(0, 0.866)$           \\ \hline
\end{tabular}
\end{table}


\section{Numerical Experiments}

\subsection{Dataset}
To conduct numeric experiments for the proposed methods, we have simulated dataset including statistics of the two-sample tests described earlier. Each observation in the dataset contains the set of computed two-sample test statistics under following parameters:
\begin{itemize} 
    \item 27 alternative hypotheses ($H_{01}-H_{09}$, $H_{11}-H_{19}$, and $H_{21}-H_{29}$);
    \item 2 competing hypotheses $S_1(t)$ and $S_2(t)$ for each alternative: 
    \begin{enumerate}
        \item if $X_1\sim\xi_1$ compares to $X_2\sim\xi_1$, then the $H_0$ is true (indicated as 'H0');
        \item if $X_1\sim\xi_1$ compares to $X_2\sim\xi_2$, then the $H_0$ is rejected (indicated as 'H1');
    \end{enumerate}    
    \item 10 sample sizes for each competing hypothesis (20, 30, 50, 75, 100, 150, 200, 300, 500, and 1000 observations);
    \item 6 censoring rates for each competing hypothesis (0\%, 10\%, 20\%, 30\%, 40\%, 50\%);
    \item 37,650 Monte Carlo replications for each competing hypothesis.
\end{itemize}

For each observation in the dataset, \textbf{the target} variable indicates 'H0' if the null hypothesis is true, or 'H1' if the null hypothesis is rejected. In total, the dataset includes 27~x~2~x~10~x~6~x~37,650~=~121,986,000 observations. We split the dataset into two samples:
\begin{enumerate}
    \item 7,650 Monte Carlo replications with all combinations of parameters ($\sim24.7$ million observations) to train the proposed methods (\textbf{Train sample});
    \item rest of 30,000 Monte Carlo replications (\textbf{Simulation sample}) for statistical investigations (test power, test statistic distributions and etc.).
\end{enumerate}

This full dataset, split into samples, source code (C++) for simulation, and a brief summary are available in the Hugging Face \href{https://huggingface.co/datasets/pfilonenko/ML_for_TwoSampleTesting}{repository} (DOI: \href{https://doi.org/10.57967/hf/2978}{10.57967/hf/2978}).

\subsection{Proposed Methods Training}

In this numeric experiment, we train the proposed methods, then test them using several metrics and assess the feature importance. To reach this, we stratify the \textbf{Train sample} into: 
\textit{train subsample} (55\%) with $\sim13.6$ million observations, 
\textit{validate subsample} (30\%) with $\sim7.4$ million observations,
and \textit{test subsample} (15\%) with $\sim3.7$ million observations.

The ratio of class balance is equal to 50\% in each sample. To find optimal hyperparameters, we apply the Optuna framework with 10 trials. The resulting metrics computed on the test sample are presented in Table~\ref{tab:training}.

\begin{table}[ht]
\centering
\caption{Training results of the proposed methods (with 95\%-th confidence intervals)}
\begin{tabular}{|lccc|}
\hline
\multicolumn{1}{|c|}{\textbf{Metric}}      & \multicolumn{1}{c|}{\textbf{LogReg}}  & \multicolumn{1}{c|}{\textbf{RF}}       & \textbf{GBM}    \\ \hline
\multicolumn{1}{|l|}{Accuracy}             & \multicolumn{1}{c|}{0.633 ± 0.0006}   & \multicolumn{1}{c|}{0.6459 ± 0.0007}   & 0.6465 ± 0.0008 \\ \hline
\multicolumn{1}{|l|}{Precision}            & \multicolumn{1}{c|}{0.6305 ± 0.0007}  & \multicolumn{1}{c|}{0.7012 ± 0.0011}   & 0.6955 ± 0.0010 \\ \hline
\multicolumn{1}{|l|}{Recall} & \multicolumn{1}{c|}{\textbf{0.6422 ± 0.0005}}  & \multicolumn{1}{c|}{0.5086 ± 0.0008}   & 0.5211 ± 0.0010 \\ \hline
\multicolumn{1}{|l|}{ROC AUC}             & \multicolumn{1}{c|}{0.686 ± 0.0009}   & \multicolumn{1}{c|}{0.7155 ± 0.0005}   & 0.7153 ± 0.0006 \\ \hline
\multicolumn{1}{|l|}{Avg. Precision}       & \multicolumn{1}{c|}{0.6945 ± 0.0011}  & \multicolumn{1}{c|}{0.743 ± 0.0006}    & 0.7423 ± 0.0007 \\ \hline
\multicolumn{1}{|l|}{Specificity}          & \multicolumn{1}{c|}{0.6237 ± 0.0010}  & \multicolumn{1}{c|}{0.7833 ± 0.0008}   & 0.7719 ± 0.0007 \\ \hline
\multicolumn{1}{|l|}{Prop. w/o FN}                & \multicolumn{1}{c|}{\textbf{0.8211 ± 0.0002}}  & \multicolumn{1}{c|}{0.7543 ± 0.0004}   & 0.7605 ± 0.0005 \\ \hline
\multicolumn{4}{|c|}{}                                                                                                                        \\ \hline
\multicolumn{1}{|c|}{\textbf{Metric}}      & \multicolumn{1}{c|}{\textbf{XGBoost}} & \multicolumn{1}{c|}{\textbf{CatBoost}} & \textbf{LAMA}   \\ \hline
\multicolumn{1}{|l|}{Accuracy}             & \multicolumn{1}{c|}{0.672 ± 0.0010}   & \multicolumn{1}{c|}{0.674 ± 0.0007}    & \textbf{0.6771 ± 0.0008} \\ \hline
\multicolumn{1}{|l|}{Precision}            & \multicolumn{1}{c|}{0.7197 ± 0.0013}  & \multicolumn{1}{c|}{0.7181 ± 0.0012}   & \textbf{0.7281 ± 0.0012} \\ \hline
\multicolumn{1}{|l|}{Recall} & \multicolumn{1}{c|}{0.5635 ± 0.0008}  & \multicolumn{1}{c|}{0.5728 ± 0.0006}   & 0.5653 ± 0.0007 \\ \hline
\multicolumn{1}{|l|}{ROC AUC}             & \multicolumn{1}{c|}{0.7478 ± 0.0009}  & \multicolumn{1}{c|}{0.7501 ± 0.0008}   & \textbf{0.7539 ± 0.0008} \\ \hline
\multicolumn{1}{|l|}{Avg. Precision}       & \multicolumn{1}{c|}{0.7772 ± 0.0008}  & \multicolumn{1}{c|}{0.7789 ± 0.0008}   & \textbf{0.7828 ± 0.0008} \\ \hline
\multicolumn{1}{|l|}{Specificity}          & \multicolumn{1}{c|}{0.7805 ± 0.0012}  & \multicolumn{1}{c|}{0.7751 ± 0.0012}   & \textbf{0.7888 ± 0.0012} \\ \hline
\multicolumn{1}{|l|}{Prop. w/o FN}                & \multicolumn{1}{c|}{0.7818 ± 0.0004}  & \multicolumn{1}{c|}{0.7864 ± 0.0003}   & 0.7827 ± 0.0003 \\ \hline
\end{tabular}
\label{tab:training}
\end{table}

The sequence of the most preferred methods based on the primary metric, Accuracy, is as follows: LAMA $>$ CatBoost $>$ XGBoost $>$ GBM or RF $>$ LogReg. This indicates that, on the \textit{test subsample}, LAMA significantly reduces Type I and Type II errors compared to the other methods.

Next, let us analyze the Proportion without False Negatives (Type II error). In this scenario, the preferred methods rank as follows: LogReg $>$ CatBoost $>$ LAMA $>$ XGBoost $>$ GBM $>$ RF. This suggests that LogReg minimizes Type II errors more effectively than the other proposed methods on the \textit{test subsample}.

Additionally, we discuss the ranking ability of the proposed methods as measured by Average Precision. This metric assesses the methods’ effectiveness in distinguishing differences in ordered values, particularly at the extremes of distributions. Given that our proposed methods focus on the right-side critical area, they demonstrate an ability to distinguish between competing hypotheses. The methods are ranked for this metric as follows: LAMA $>$ CatBoost $>$ XGBoost $>$ RF or GBM $>$ LogReg.

We also evaluate the importance of the features included in the final versions of each proposed method. For this, we use two methods: Permutation Importance (available for all methods) and Feature Importance (available for some methods, as an additional assessment). The results are shown in Tables~\ref{tab:PI} and~~\ref{tab:FI}.

\begin{table}[]
\centering
\caption{Permutation Importance (features sorted by average rank among all methods)}
\begin{tabular}{|l|c|c|c|c|c|c|}
\hline
\multicolumn{1}{|c|}{\textbf{feature}} & \textbf{LogReg} & \textbf{RF} & \textbf{GBM} & \textbf{XGBoost} & \textbf{CatBoost} & \textbf{LAMA} \\ \hline
BN-SCE                          & 21.24           & 0.95        & 1.64        & 18.92            & 26.81             & 12.03         \\ \hline
BN-GPH                          & 4.42            & 0.18        & 0.30        & 24.20            & 23.53             & 11.49         \\ \hline
Tarone-Ware                  & 1.01            & 0.15        & 0.05        & 27.66            & 26.75             & 12.72         \\ \hline
MIN3                             & 1.02            & 3.33        & 2.55        & 16.16            & 12.56             & 7.84          \\ \hline
BN-MCE                          & 15.59           & 2.15        & 3.59        & 6.58             & 7.64              & 4.41          \\ \hline
WKM                              & 8.22            & 0.06        & 0.02        & 21.37            & 16.55             & 10.70         \\ \hline
Gehan                            & 21.36           & 0.05        & 0.03        & 13.30            & 17.83             & 8.33          \\ \hline
MAX                       & 2.58            & 0.06        & 0.02        & 12.85            & 16.93             & 9.58          \\ \hline
log-rank                          & 0.08            & 0.13        & 0.15        & 12.81            & 13.20             & 7.00          \\ \hline
Peto                             & 1.59            & 0.04        & 0.00        & 17.40            & 7.21              & 8.70          \\ \hline
Peto-Prentice                & 0.21            & 0.08        & 0.13        & 7.70             & 9.47              & 4.31          \\ \hline
Prentice                    & 1.29            & 0.10        & 0.07        & 2.38             & 3.76              & 3.12          \\ \hline
$min(n_1,n_2) \ge 500$                           & 0.42           & 0.99        & 1.09        & 1.95             & 1.66              & 0.77          \\ \hline
$Q$-test                                & 0.53            & 0.03        & 0.04        & 5.83             & 4.28              & 4.91          \\ \hline
$0 < min(n_1,n_2) < 50$                          & 0.08           & 0.48        & 0.44        & 0.29             & 0.46              & 0.20          \\ \hline
$100 \le min(n_1,n_2) < 500$                       & 0.42            & 0.08        & 0.05        & 0.52             & 0.39              & 0.09          \\ \hline
$50 \le min(n_1,n_2) < 100$                        & 0.08            & 0.10        & 0.08        & 0.06             & 0.12              & 0.05          \\ \hline
$\frac{n_1r_1 + n_2r_2}{n_1+n_2} > 35\%$                       & 0.68            & 0.00        & 0.00        & 0.27             & 0.35              & 0.24          \\ \hline
$\frac{n_1r_1 + n_2r_2}{n_1+n_2} = 0\%$                       & 0.04            & 0.00        & 0.00        & 0.19             & 0.25              & 0.21          \\ \hline
$15\% < \frac{n_1r_1 + n_2r_2}{n_1+n_2} \le 35\%$                        & 0.40            & 0.00        & 0.00        & 0.05             & 0.03              & 0.03          \\ \hline
$0\% < \frac{n_1r_1 + n_2r_2}{n_1+n_2} \le 15\%$                        & 0.11            & 0.00        & 0.00        & 0.03             & 0.05              & 0.05          \\ \hline
\end{tabular}
\label{tab:PI}
\end{table}

\begin{table}[]
\centering
\caption{Feature Importance (features sorted by average rank among all methods)}
\begin{tabular}{|l|c|c|c|c|}
\hline
\multicolumn{1}{|c|}{\textbf{feature}}            & \textbf{CatBoost} & \textbf{XGBoost} & \textbf{RF} & \textbf{GBM} \\ \hline
MIN3                                              & 20.08             & 40.16            & 24.37       & 17.81       \\ \hline
$min(n_1,n_2) \ge 500$                            & 12.37             & 18.18            & 7.99        & 9.23        \\ \hline
BN-MCE                                            & 7.33              & 4.43             & 20.44       & 29.32       \\ \hline
BN-SCE                                            & 10.48             & 3.46             & 13.90       & 12.20       \\ \hline
BN-GPH                                            & 8.88              & 2.07             & 10.01       & 15.50       \\ \hline
$0 < min(n_1,n_2) < 50$                           & 1.30              & 4.59             & 2.53        & 3.71        \\ \hline
Tarone-Ware                                       & 8.33              & 1.57             & 1.46        & 0.87        \\ \hline
log-rank                                          & 7.44              & 1.84             & 0.99        & 0.94        \\ \hline
Gehan                                             & 2.68              & 1.04             & 3.17        & 1.27        \\ \hline
$100 \le min(n_1,n_2) < 500$                      & 1.86              & 9.16             & 1.14        & 0.85        \\ \hline
$Q$-test                                          & 4.51              & 1.35             & 0.70        & 4.35        \\ \hline
WKM                                               & 3.56              & 1.41             & 2.62        & 0.60        \\ \hline
Peto                                              & 1.47              & 0.82             & 2.59        & 1.03        \\ \hline
MAX                                               & 2.00              & 1.02             & 5.28        & 0.40        \\ \hline
$50 \le min(n_1,n_2) < 100$                       & 0.35              & 2.81             & 0.96        & 0.78        \\ \hline
Prentice                                          & 2.39              & 0.84             & 1.02        & 0.52        \\ \hline
Peto-Prentice                                     & 2.94              & 0.62             & 0.83        & 0.64        \\ \hline
$\frac{n_1r_1 + n_2r_2}{n_1+n_2} > 35\%$          & 1.05              & 1.55             & 0.00        & 0.00        \\ \hline
$\frac{n_1r_1 + n_2r_2}{n_1+n_2} = 0\%$           & 0.74              & 1.60             & 0.00        & 0.00        \\ \hline
$0\% < \frac{n_1r_1 + n_2r_2}{n_1+n_2} \le 15\%$  & 0.15              & 0.69             & 0.00        & 0.00        \\ \hline
$15\% < \frac{n_1r_1 + n_2r_2}{n_1+n_2} \le 35\%$ & 0.10              & 0.78             & 0.00        & 0.00        \\ \hline
\end{tabular}
\label{tab:FI}
\end{table}

As a result, we present several key findings for the proposed methods, including: a)~as expected in~\cite{MIN3}, one of most significant features are the two-sample Bagdonavičius-Nikulin and MIN3 tests; b) features based on the sample size are more significant than features based on the censoring rate; c) features of the censoring rate are at a level close to the insignificance.

\subsection{Study of the Test Power}

In this experiment, we determine the statistical power of the proposed methods. To do this, we compute estimates of the power of all the methods considered using the alternative hypotheses described earlier. As the result obtained, rather than reporting absolute power values, we compute the \textit{average rank} among the set of methods considered, since the average rank more clearly reflects the preference of one method in comparison with rest methods. Detailed absolute values of the test power are represented in Supplementary materials.

For values of the average rank, we also compute the scores of the Wald~\cite{wald} and Savage~\cite{savage} criteria, which are used for decision-making under risk and uncertainty of environment. In this case, the "environment" is an alternative hypothesis, and \textit{the utility function} is the rank of the method on the alternative hypothesis. The lower the rank of the method, the higher it is in the ordered list, which means that this method is better than others. The Wald criterion focuses on maximizing utility in worst-case scenarios, whereas the Savage criterion aims to minimize 'regret' associated with the utility loss compared to the best possible outcome. The same methodology has already applied previously in~\cite{MIN3}.

\begin{figure}[ht]
    \centering
    \includegraphics[height=22.5cm]{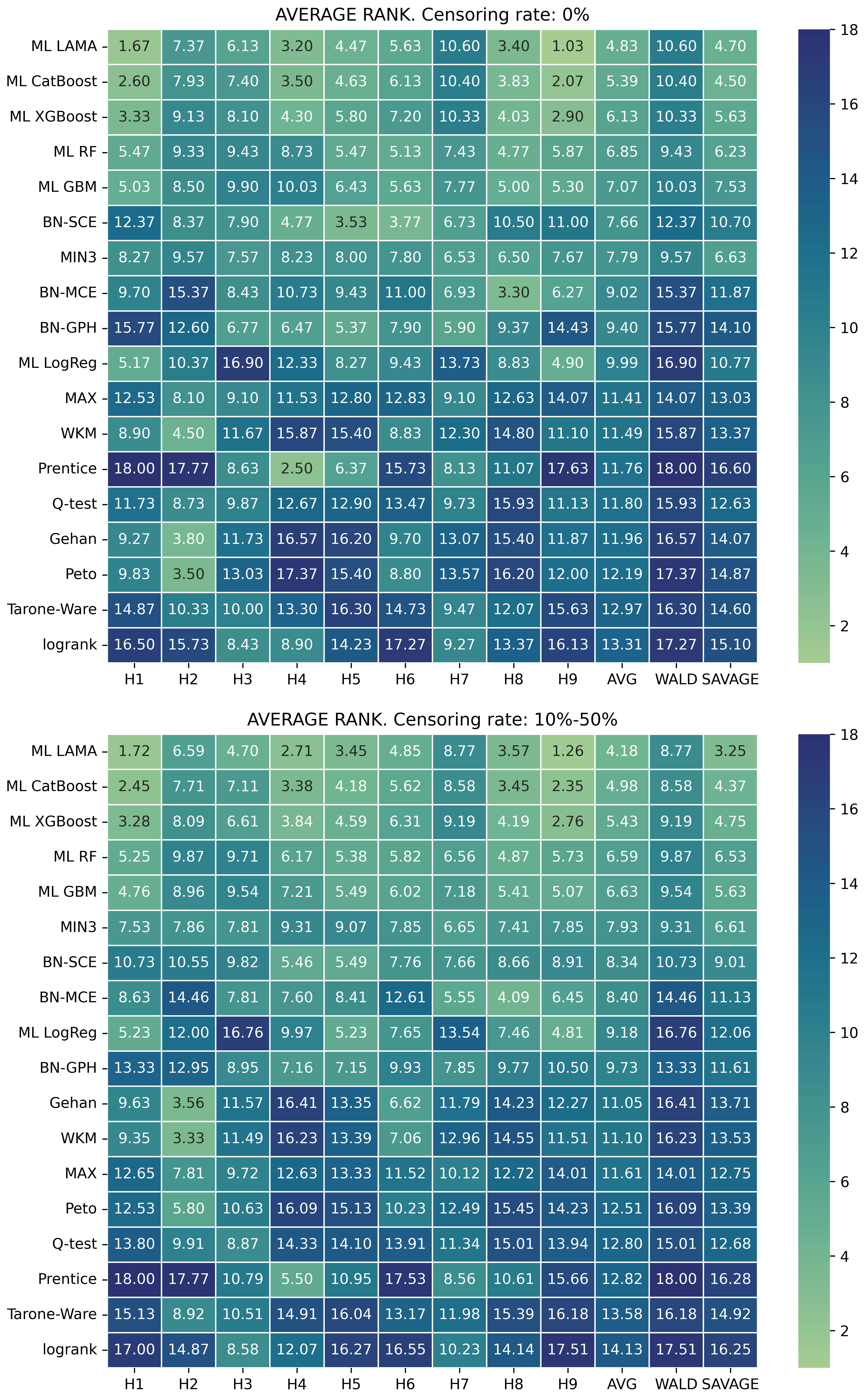}
    \caption{Average rank (AVG) of tests including the Wald and Savage criteria on the groups $H_{\text{I}}-H_{\text{IX}}$}
    \label{fig:avg_rank}
\end{figure}

To reach this, we provide the investigation using the \textbf{Simulation sample} for each alternative group in the range $H_I$-$H_{IX}$ under censoring rates 0\% and 0\%-50\%. We present results in Figure~\ref{fig:avg_rank}. The obtained results are ordered by the average rank values (AVG column). 

The results indicate that the proposed methods are at the top of the ordered lists (under censoring rate 0\%-50\%) with the exception of LogReg. This suggests that the proposed methods, on average, have a more preferable test power than the classic two-sample tests in terms of average rank. Comparing the values of the Wald criterion (assuming that the environment is in the worst state), we note that the proposed methods are also more preferable (under censoring rate 0\%-50\%). Nevertheless, the classic two-sample MIN3 test was among the TOP-4 preferred methods, since the construction of this test statistic is based on the same assumption. Comparing the values of the Savage criterion ('regret' minimization), we note that the proposed methods also demonstrate superior performance (under censoring rate 0\%-50\%). Thus, we have shown that the proposed methods have high potential, since they have higher test power in the numerical experiments conducted.

\subsection{Study of the Test Statistic Distribution under the $H_0$}
Another investigation that is necessary for practical application is the study of the test statistic distribution. This ensures the accurate determination of the critical value in hypothesis testing. As is known, the actual (pre-limit) distribution of statistics can differ from the limit distribution and depend on many factors, such as sample size,  censoring rate, data distribution, and many others. Therefore, it is important to evaluate its behavior in a numerical experiment in order to make correct statistical conclusions in practice.

The design of our numeric experiment is as follows:
\begin{enumerate}
    \item  For each set of parameters $(n_1, n_2, r_1, r_2, F, F^C)$, we plot the CDF of the proposed method under the $H_0$ is true. We briefly denote such a distribution as $G(S|H_0)$. Each $G(S|H_0)$ contains 30,000 observations of the \textbf{Simulation sample};
    
    \item Construct a minorant $G_{MIN}(S|H_0)$, a majorant $G_{MAX}(S|H_0)$ among the set of the CDFs (obtained in item~1), and the middle line between them: $G_{AVG}(S|H_0) = (G_{MIN}(S|H_0) + G_{MAX}(S|H_0)) / 2$;

    \item Present the average deviation of the middle line from the minorant/majorant in the right tail with probabilities $p$ in the range [0.9, 1.0). This is related to the fact that the proposed methods have right-critical area. Hence, it corresponds to area of the test size $0 < \alpha \le 0.1$).
\end{enumerate}

\begin{landscape}
    \begin{figure}[ht]
    \centering
    \includegraphics[height=\textheight]{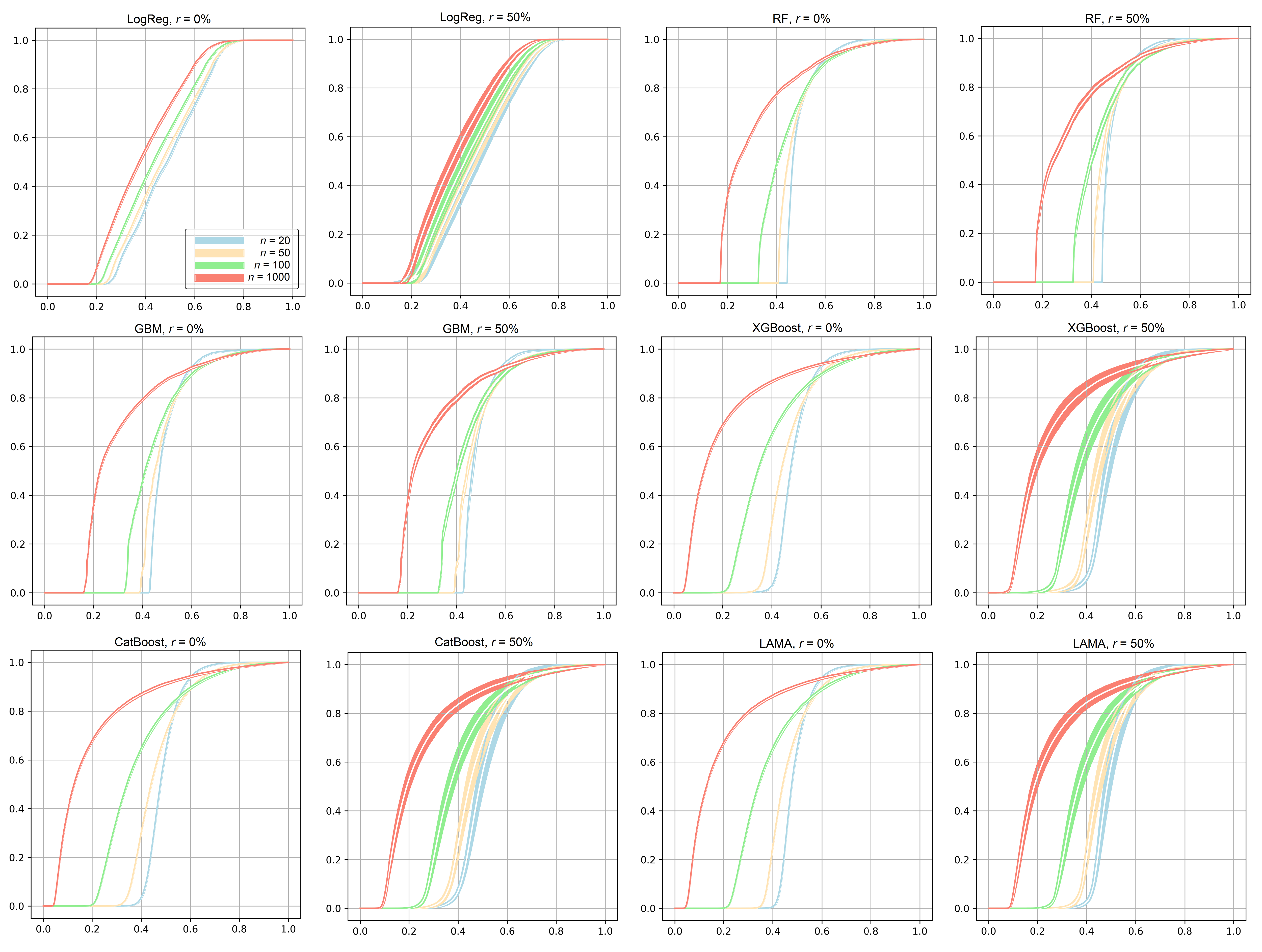}
    \caption{$G(S|H_0)$ distributions for the proposed methods. White line is the middle line. Color stripe is the region between minorant and majorant.}
    \label{fig:GSH0}
\end{figure}
\end{landscape}

This way allow us to estimate the average error in the $p$-value computation and assess a confidence interval for the $p$-value when solving a practical task. Table~\ref{tab:GSH0} presents computed average errors varing the proposed method, sample size, and censoring rate. Figure~\ref{fig:GSH0} shows the same results graphically.

\begin{table}[ht]
\centering
\caption{Average error of the $p$-value computation simulated for the test size $0 \le \alpha \le 0.1, H_{01}-H_{29}$}
\begin{tabular}{|c|c|c|c|c|c|c|c|}
\hline
\textbf{$r_i$}        & \textbf{$n_i$} & \textbf{LogReg} & \textbf{RF} & \textbf{GBM} & \textbf{XGBoost} & \textbf{CatBoost} & \textbf{LAMA} \\ \hline
\multirow{4}{*}{0\%}  & 20             & 0.0053          & 0.0141      & 0.0136       & 0.0287           & 0.0231            & 0.0241        \\ \cline{2-8} 
                      & 50             & 0.0079          & 0.0145      & 0.0084       & 0.0149           & 0.0166            & 0.0189        \\ \cline{2-8} 
                      & 100            & 0.0084          & 0.0112      & 0.0078       & 0.0087           & 0.0088            & 0.0106        \\ \cline{2-8} 
                      & 1000           & 0.0111          & 0.0100      & 0.0080       & 0.0069           & 0.0090            & 0.0083        \\ \hline
\multirow{4}{*}{10\%} & 20             & 0.0077          & 0.0140      & 0.0114       & 0.0224           & 0.0284            & 0.0236        \\ \cline{2-8} 
                      & 50             & 0.0072          & 0.0177      & 0.0099       & 0.0212           & 0.0218            & 0.0251        \\ \cline{2-8} 
                      & 100            & 0.0091          & 0.0127      & 0.0086       & 0.0112           & 0.0114            & 0.0100        \\ \cline{2-8} 
                      & 1000           & 0.0157          & 0.0122      & 0.0091       & 0.0170           & 0.0160            & 0.0133        \\ \hline
\multirow{4}{*}{20\%} & 20             & 0.0105          & 0.0170      & 0.0159       & 0.0229           & 0.0224            & 0.0203        \\ \cline{2-8} 
                      & 50             & 0.0057          & 0.0158      & 0.0096       & 0.0173           & 0.0209            & 0.0171        \\ \cline{2-8} 
                      & 100            & 0.0075          & 0.0111      & 0.0092       & 0.0124           & 0.0148            & 0.0093        \\ \cline{2-8} 
                      & 1000           & 0.0166          & 0.0107      & 0.0086       & 0.0177           & 0.0166            & 0.0099        \\ \hline
\multirow{4}{*}{30\%} & 20             & 0.0105          & 0.0175      & 0.0164       & 0.0283           & 0.0319            & 0.0315        \\ \cline{2-8} 
                      & 50             & 0.0055          & 0.0178      & 0.0112       & 0.0195           & 0.0206            & 0.0163        \\ \cline{2-8} 
                      & 100            & 0.0090          & 0.0161      & 0.0150       & 0.0218           & 0.0245            & 0.0165        \\ \cline{2-8} 
                      & 1000           & 0.0144          & 0.0122      & 0.0107       & 0.0176           & 0.0206            & 0.0105        \\ \hline
\multirow{4}{*}{40\%} & 20             & 0.0122          & 0.0201      & 0.0189       & 0.0499           & 0.0521            & 0.0484        \\ \cline{2-8} 
                      & 50             & 0.0070          & 0.0145      & 0.0114       & 0.0197           & 0.0223            & 0.0155        \\ \cline{2-8} 
                      & 100            & 0.0074          & 0.0155      & 0.0138       & 0.0128           & 0.0186            & 0.0144        \\ \cline{2-8} 
                      & 1000           & 0.0138          & 0.0114      & 0.0111       & 0.0205           & 0.0279            & 0.0128        \\ \hline
\multirow{4}{*}{50\%} & 20             & 0.0160          & 0.0240      & 0.0231       & 0.0791           & 0.0882            & 0.0873        \\ \cline{2-8} 
                      & 50             & 0.0064          & 0.0169      & 0.0123       & 0.0399           & 0.0421            & 0.0327        \\ \cline{2-8} 
                      & 100            & 0.0081          & 0.0136      & 0.0117       & 0.0196           & 0.0225            & 0.0158        \\ \cline{2-8} 
                      & 1000           & 0.0100          & 0.0115      & 0.0103       & 0.0339           & 0.0334            & 0.0216        \\ \hline
\end{tabular}
\label{tab:GSH0}
\end{table}

The results indicate that cumulative distribution functions (CDFs) have the largest distance $|G_{MAX}(S|H_0)-G_{MIN}(S|H_0)|$ in the central area, and a significant decrease in distance occurs towards the tails of the distributions. Note that the proposed methods LogReg, RF and GBM have a smaller range between CDFs than XGBoost, CatBoost and LAMA. However, this is almost leveled out already when the sample size close to 100 observations. The deviations calculated in Table 5 suggest that the Monte Carlo simulation effectively addresses the determination of critical values in hypothesis testing. Thus, we have prepared percentage points which can be calculated for the proposed methods in order to use them in practice. Detailed values of percentage points are presented in Supplementary materials.

\section{Discussion}

Thus, in this paper we proposed methods for two-sample testing under right-censored data based on various machine learning methods.

To train and test the proposed methods, we simulated a synthetic dataset (the dataset and source code are available in GitHub) containing over 121 million objects.

We trained the proposed methods on the training part of the dataset, computed metrics and determined which methods make the least number of type I and type II errors on the test part of the data (LAMA~$>$ CatBoost~$>$ XGBoost~$>$ GBM or RF~$>$ LogReg). The significance of features for the proposed methods based on Feature Importance and Permutation Importance was shown.

A power study was conducted for the proposed methods. On a set of close alternatives, we averaged the ranks of the methods, sorting them by test power. The lower the average rank, the more preferable the method is, on average. It was found that the proposed methods (especially LAMA and CatBoost) are more preferable methods for two-sample testing at a censoring rate of 0\%-50\%.

The behavior of the test statistic distribution of the proposed methods under the $H_0$ is true is also investigated. For this purpose, we simulated $G(S|H_0)$ under various parameters $(n_i,r_i,F_i,F^C_i),i=1,2$ and computed the average error between the middle line and the minorant/majorant of this set (in the region $0 < \alpha \le 0.1$). This shows what order of error can be expected when calculating the $p$-value. The obtained results show that Monte Carlo simulating the test statistic distribution to determine the critical value can be an effective way to solve this issue. A detailed table of percentage points was prepared.

The advantage of the proposed methods is that it is based on many classical and well-studied two-sample tests. This allows the method to be sensitive to the properties of these tests and aggregate their synergistic effect. Similar approaches based on the selection function have been used previously in~\cite{q_test,MAX,MIN3}. The choice of such a function was associated with taking into account the special statistical properties of the methods, which was a limitation for a wide study in this direction. ML methods allow one to be more sensitive to the properties of the underlying methods and at the same time play the role of the \textit{selection function}. Note also that the obtained test power results confirm the results of when studying the MIN3 test~\cite{MIN3}.

To apply these methods in practice, the full source code for the implementation of the proposed methods is published in the GitHub repository (including trained models and scripts to run). If even this is not enough, the same repository contains the full source code for modeling the original dataset.

The proposed methods can find its application in solving problems in a variety of fields: medicine, biology, sociology, engineering problems and many others, where a comparison of two samples of right-censored observations (as well as complete observations) is required. However, for example, in genetics problems, the application of the proposed methods can be complicated by the lack of the limit distribution in an analytical form for calculating $p$-value at very low values of the type I error (e.g., $\alpha < 0.001$). Nevertheless, this issue can also be resolved by increasing the volume of the Monte Carlo simulation.

However, one should not forget about the \textit{limitations} of this work, which are as follows:
(1) $n_1 = n_2$ and $r_1 = r_2$. It is necessary to check how the proposed methods behave under conditions of unequal sample sizes and/or censoring rates;
(2) $n_1,n_2 \le 1000$. Often, classical two-sample tests already have high power at such sample sizes. However, the behavior of the proposed methods for sample sizes greater than 1000 observations has not been investigated;
(3) $r_1,r_2 \le 50\%$. The extremely high level of censoring rate ($r > 50\%$) requires additional investigation;
(4) The set of alternative hypothesis has been various but still limited.

The study the limit null distribution of the proposed methods can significantly increase the attractiveness and accessibility of these methods for solving practical problems. In our opinion, this can be an one of the prioritiest task for further possible research.

To summarize all of the above, we note that the use of the machine learning paradigm has already allowed us to qualitatively improve the solution of various applied problems. And this study also shows that the use of machine learning methods can significantly improve the accuracy of statistical inferences when two-sample problem testing.

To conduct the study, the following hardware is utilized: Intel® Core™ i7-12700H, 64GB RAM, 1TB ROM, and NVIDIA GeForce RTX 3060 Mobile.


\section{Conclusion}

In this paper, we demonstrated that the application of machine learning methods to solve the two-sample problem with right-censored observations can significantly enhance the accuracy of statistical inferences. To achieve this, we formulated the solution to the problem as a stacking of classical two-sample tests using machine learning methods as a selection function.

We investigated the statistical power and behavior of the null distribution of the proposed methods. To conduct our numerical experiments, we prepared a synthetic dataset using Monte Carlo simulations. Despite the absence of the limit distribution necessary for the $p$-value computation (this issue can be resolved using Monte Carlo simulation), the results indicate that the proposed methods improve the accuracy of statistical inferences in two-sample testing, whether under right-censored data or not.

The results obtained can find applications in many fields of science (engineering, chemistry, physics, biology, medicine, etc.) where comparison of two samples with right-censored (incomplete) observations is required. The use of the proposed methods will enhance the accuracy of decision-making in these applied areas.

To be able to reproduce the results or apply the proposed methods, we have published the original synthetic dataset, including the source code (Hugging Face), and trained models, including example scripts (GitHub).


\section*{Supplemental Materials}
The supplementary materials contain:
(1) Table with absolute power values;
(2) Table with percentage points for the proposed methods.


\section*{Data Availability Statement}

The dataset simulated during and analyzed during the current study is available in the Hugging Face \href{https://huggingface.co/datasets/pfilonenko/ML_for_TwoSampleTesting}{repository}, DOI:~\href{https://doi.org/10.57967/hf/2978}{10.57967/hf/2978}.

The source code of the proposed methods is available in the GitHub repository: \\
\href{https://github.com/pfilonenko/ML_for_TwoSampleTesting}{\textit{https://github.com/pfilonenko/ML\_for\_TwoSampleTesting}}.


\section*{Disclosure Statement}
The authors have no conflicts of interest to declare.


\section*{Funding}
This research received no specific grant from funding agencies in the public, commercial, or not-for-profit sectors.





\end{document}